\documentclass{article}

\usepackage[numbers]{natbib}
\usepackage[preprint]{neurips_2026}

\usepackage[utf8]{inputenc} 
\usepackage[T1]{fontenc}    
\usepackage{hyperref}       
\usepackage{url}            
\usepackage{booktabs}       
\usepackage{amsfonts}       
\usepackage{nicefrac}       
\usepackage{microtype}      
\usepackage{xcolor}         

\usepackage{graphicx}
\usepackage{animate}
\usepackage{amsmath}
\usepackage{amssymb}
\usepackage{multirow}

\title{StreamingEffect: Real-Time Human-Centric Video Effect Generation}

\author{
  Yiren Song\thanks{Equal contribution.} \quad
  Cheng Liu\footnotemark[1] \quad
  Yuxin Jiang \quad
  Mike Zheng Shou\thanks{Corresponding author.} \\
  \\
  Show Lab, National University of Singapore
}

\begin{document}

\maketitle
\vspace{-6mm}

\begin{figure*}[h]
\animategraphics[width=\linewidth]{5}{figs/Teaser2/frame_}{01}{09}
\vspace{-8mm}
\caption{Given an incoming human-centric video stream, StreamingEffect supports interactive
real-time control by switching the reference image or text prompt on the fly, enabling stable
long-term effects without flickering or drift. Readers can click and play the video clips using
{\color{red}\textbf{Adobe Acrobat}}.}
\label{fig:Teaser}
\end{figure*}

\begin{abstract}
Streaming video effect generation is highly desirable for live human-centric applications such as e-commerce streaming, entertainment, and vlogging, yet remains difficult due to the lack of suitable data and deployable editing models. Unlike generic video generation, this task requires real-time video-to-video editing that adds expressive effects while preserving human identity, background content, and temporal consistency. Existing acceleration efforts mainly focus on text-to-video generation, while efficient distillation for video editing remains largely underexplored. In this paper, we present \textbf{StreamingEffect}, a real-time human-centric streaming video effect framework. We adopt an in-context video editing architecture and train a high-quality bidirectional teacher, then distill it into a causal autoregressive student and further reduce sampling from 50 steps to 4 steps. We also introduce keyframe control, allowing reference effect frames to be injected online and propagated through the stream for interactive editing. To address the data bottleneck, we construct \textbf{VideoEffect-130K}, to our knowledge the largest human-centric video effect dataset, containing 70K effect videos and 60K editing videos across 600 effect categories curated from short-video and editing platforms. Experiments show that our method enables real-time, high-quality 720p video editing on a single H200 GPU. Code is released at \href{https://github.com/showlab/StreamingEffect}{https://github.com/showlab/StreamingEffect}
\end{abstract}

\section{Introduction}

Live video has become a dominant medium for digital interaction, spanning e-commerce streaming, short-video creation, entertainment, and personal vlogging. 
In these scenarios, visual effects play an important role in improving attractiveness and engagement, by adding accessories, stylized appearances, decorative elements, or atmosphere cues to human-centric videos. 
Although such effects are widely desired in real applications, current industry systems still largely rely on manually designed assets and detection-tracking-attachment pipelines. 
These template-based solutions are efficient for predefined effects, but are limited in diversity, semantic flexibility, and visual expressiveness, making it difficult to support open-ended and instruction-driven effect creation.

Streaming video effect generation is therefore a highly demanded but technically challenging problem. 
Given an incoming video stream, the goal is to continuously generate edited frames with expressive and controllable effects under strict latency constraints. 
This task differs from conventional offline video editing in three key aspects. 
First, it requires real-time streaming inference rather than full-sequence generation. 
Second, it must preserve editing consistency: the human identity, body structure, background, and scene content should remain stable before and after editing, while only the desired effects are added. 
Third, it requires flexible controllability, allowing effects to be specified not only by text but also by reference images or keyframes for interactive creation.

Despite its practical importance, this problem remains underexplored due to two major gaps. 
The first is a data gap. 
Existing video editing datasets~\cite{lin2026kiwi, wu2025insvie, yu2025veggie} are often small-scale, synthetic, or not tailored to human-centric effect generation. 
In particular, there is a lack of large-scale paired real-video data that captures how diverse visual effects are applied to faces, bodies, clothing, and surrounding human-centric scenes. 
To address this gap, we construct \textbf{VideoEffect-130K}, a large-scale paired dataset built from both platform-style effect sources and synthetic video editing pipelines, covering real short-video and editing platforms such as TikTok, Douyin, Kuaishou, and Jianying-style effect sources.
It contains 130K high-quality paired videos covering 600 effect categories, and is, to our knowledge, the largest dataset specifically designed for human-centric video effect generation.

The second gap lies in efficient video editing deployment.
Recent acceleration techniques have made substantial progress in text-to-video generation, including causal generation~\citep{yin2024causvid,huang2025selfforcing,lin2025apt} and few-step distillation~\citep{salimans2022progressive,song2023consistency,yin2024dmd}.
However, streaming video editing is fundamentally different: the model must preserve the input video content, follow effect conditions, maintain temporal consistency, and generate edited frames with low latency.
Efficient distillation for video editing, especially for human-centric effect generation, remains far less explored.
Meanwhile, strong image editing models~\citep{brooks2023instructpix2pix,zhang2023controlnet,kawar2023imagic,hertz2023ptp} can produce high-quality effect keyframes, but they operate on individual images and cannot directly propagate effects through a video stream.

In this paper, we present \textbf{StreamingEffect}, a real-time streaming video-to-video framework for human-centric effect generation. 
We start from an in-context video editing architecture, where the input video and effect conditions are jointly modeled in a shared context space. 
This design allows the model to preserve source-video information while adding new effects, which is essential for maintaining identity, background, and temporal consistency. 
We first train a high-quality bidirectional teacher model on short clips to obtain strong visual quality and temporal coherence. 
To make the model deployable in streaming scenarios, we adopt a two-stage distillation pipeline inspired by recent advances in causal video distillation~\citep{yin2024causvid,huang2025selfforcing}: bidirectional-to-causal autoregressive distillation for online chunk prediction, followed by multi-step-to-few-step distillation~\citep{song2023consistency,yin2024dmd,sauer2023add} that reduces sampling from 50 steps to 4 steps.

To further improve controllability and visual quality, we introduce keyframe control. 
Instead of relying only on text prompts, users or upstream image editing models can provide a reference effect keyframe during streaming inference. 
StreamingEffect then propagates the reference effect to subsequent frames in a temporally coherent manner, enabling image-driven effect control and interactive online updates. 
With system-level optimization, our framework achieves real-time, high-quality 720p video editing on a single H200 GPU.

Our contributions are summarized in three aspects:
\begin{itemize}
    \item We introduce \textbf{streaming video effect generation}, a practical yet challenging task for live human-centric applications, requiring real-time inference, editing consistency, aesthetic quality, and flexible effect control.

    \item We propose \textbf{StreamingEffect}, a real-time streaming video-to-video framework that integrates in-context video editing, causal autoregressive distillation, few-step acceleration, and keyframe-based interactive control.

    \item We build \textbf{VideoEffect-130K}, a 130K-pair dataset covering 600 effect and editing categories, combining rendered effect data with synthetic/general-editing data for controllable human-centric video effect generation.
\end{itemize}

\section{Related Work}

\paragraph{Video Generation Models.}

\paragraph{Video Generation Models.}
Recent video generation has been driven by diffusion- and transformer-based foundation models built on image diffusion priors~\citep{ho2020ddpm,song2021ddim,rombach2022ldm,peebles2023dit}.
Early landmark video diffusion systems including \emph{VDM}~\citep{ho2022vdm}, \emph{Imagen Video}~\citep{ho2022imagenvideo}, \emph{Make-A-Video}~\citep{singer2023makeavideo}, \emph{Video LDM}~\citep{blattmann2023videoldm}, \emph{AnimateDiff}~\citep{guo2024animatediff}, and \emph{Lumiere}~\citep{bartal2024lumiere} establish the recipe of jointly modeling spatial and temporal context, while commercial-scale systems such as \emph{Sora}~\citep{openai2024sora} demonstrate emerging large-scale capability.
\emph{Stable Video Diffusion} further shows that scaling latent video diffusion with curated data produces strong text-to-video and image-to-video synthesis priors~\citep{blattmann2023stable}, and recent open systems including \emph{Latte}~\citep{ma2024latte}, \emph{Emu Video}~\citep{girdhar2024emuvideo}, \emph{VideoCrafter2}~\citep{chen2024videocrafter2}, \emph{CogVideoX}~\citep{yang2025cogvideox}, \emph{HunyuanVideo}~\citep{kong2024hunyuanvideo}, and \emph{Wan}~\citep{wan2025wan} continue to push motion quality, semantic alignment, duration, and fidelity through stronger spatiotemporal representations, text-video fusion, and large-scale training recipes.
Beyond unconditional or text-driven synthesis, controllable video generation further incorporates visual, structural, motion, trajectory, layout, or reference-based conditions to improve user control over generated content~\citep{jiang2025vace, zhang2025easycontrol, song2025worldwander, song2024processpainter, ma2024followpose, ma2025followcreation, ma2024followyouremoji, ma2025followyourclick, song2025mitty, yang2025x, song2026omnihumanoid}.
These models have greatly advanced open-source video generation, but they are primarily designed for offline creation where quality is prioritized over low-latency streaming.
Thus, they provide powerful generative backbones but do not directly address real-time, human-centric video effect editing.

\paragraph{Real-Time Video Generation.}
Video diffusion models remain costly due to large model sizes, full-sequence attention, and multi-step denoising.
Building on a long line of step-distillation methods originally developed for image diffusion---progressive distillation~\citep{salimans2022progressive}, consistency models~\citep{song2023consistency,luo2023lcm}, distribution-matching distillation~\citep{yin2024dmd,yin2024dmd2}, adversarial diffusion distillation~\citep{sauer2023add}, rectified flow~\citep{liu2023rectifiedflow}, and one-step generators~\citep{liu2024instaflow}---recent work pushes video efficiency through compressed latent modeling, streaming architectures, and distillation.
\emph{LTX-Video} operates in highly compressed latent spaces, while \emph{StreamDiT} introduces streaming generation with moving buffers~\citep{kodaira2025streamdit}.
Distillation-based methods including \emph{DOLLAR}~\citep{ding2024dollar}, 
\emph{Diagonal Distillation}~\citep{liu2026diagonal}, 
\emph{Rolling Forcing}~\citep{liu2025rolling}, 
\emph{CausVid}~\citep{yin2024causvid}, 
\emph{Self-Forcing}~\citep{huang2025selfforcing}, 
and \emph{APT}~\citep{lin2025apt} reduce sampling steps or convert 
bidirectional/diffusion generators into causal or long-horizon autoregressive models.
These works highlight the importance of combining autoregressive streaming with step reduction, but mainly target generic video generation rather than controllable human-centric video-to-video editing, where source preservation, effect consistency, and interactive controllability are equally critical.

\paragraph{Video Editing and Visual Effects.}
Video editing has progressed from tuning-free adaptations of image diffusion models~\citep{wu2023tuneavideo,qi2023fatezero,ceylan2023pix2video,liu2024videop2p,cheng2023insv2v, song2025omniconsistency, guo2025any2anytryon, zhang2024ssr, wang2025diffdecompose, gong2025relationadapter} to video-native editing frameworks.
\emph{Rerender A Video}, \emph{TokenFlow}, and \emph{CoDeF} exploit image diffusion priors with temporal correspondence, feature propagation, or canonical-field representations for coherent text-guided editing~\citep{yang2023rerender,geyer2023tokenflow,ouyang2024codef, song2024processpainter}.
\emph{AnyV2V} decomposes video editing into first-frame image editing---enabled by strong image editors such as \emph{InstructPix2Pix}~\citep{brooks2023instructpix2pix}, \emph{ControlNet}~\citep{zhang2023controlnet}, \emph{Imagic}~\citep{kawar2023imagic}, and \emph{Prompt-to-Prompt}~\citep{hertz2023ptp}---followed by image-to-video generation, while \emph{VACE} unifies video creation and editing through a video-native framework~\citep{ku2024anyv2v,jiang2025vace}.
\emph{StreamV2V} studies online video-to-video translation with feature banks, but focuses on general prompt-based translation rather than effect-centric human video editing~\citep{liang2024streamv2v}~\citep{lin2026kiwi}~\citep{pan2026omniweaving}.
For visual effects, \emph{VFX Creator}~\citep{liu2025vfxcreator}, 
\emph{VFXMaster}~\citep{li2025vfxmaster}, 
\emph{Transanimate}~\citep{chen2025transanimate}, 
and \emph{IC-Effect}~\citep{li2025iceffect} explore controllable VFX generation 
with mask control, reference videos, or in-context conditioning.
In contrast, our work targets real-time human-centric streaming video effect generation, emphasizing low-latency deployment, source-video preservation, and online keyframe control.

\section{Method}
\label{sec:method}

\begin{figure*}[!t]
\centering
\includegraphics[width=\linewidth]{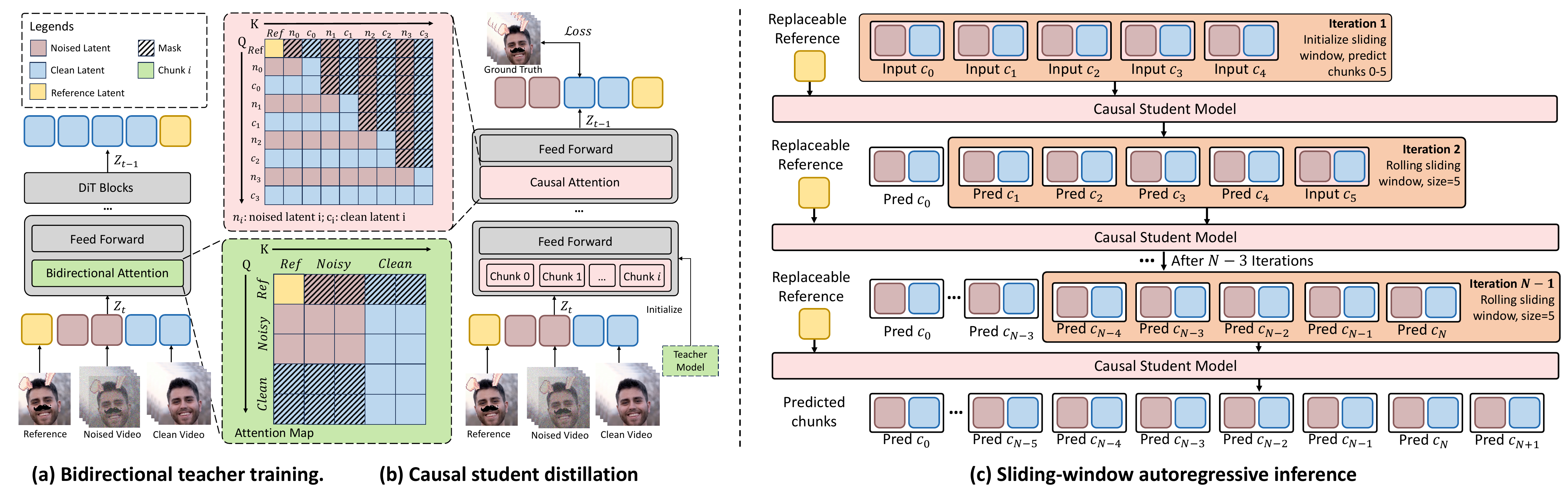}
\caption{\textbf{Overview of StreamingEffect.} (a) \textbf{Bidirectional Teacher Training:} A bidirectional teacher is trained with reference-conditioned in-context video editing. (b) \textbf{Causal Student Distillation:} The teacher is distilled into a causal autoregressive student for streaming generation. (c) \textbf{Sliding-window Autoregressive Inference:} The student edits incoming video chunks online with cached context and propagates effects through the stream.}
\label{fig:method}
\end{figure*}

We present \textbf{StreamingEffect}, a real-time streaming video-to-video framework for human-centric effect generation. 
Sec.~\ref{sec:problem} defines the streaming formulation, Sec.~\ref{sec:architecture} introduces our reference-conditioned in-context editing architecture, Sec.~\ref{sec:teacher} trains a high-quality bidirectional teacher, and Sec.~\ref{sec:distillation} converts it into a causal few-step streaming student. 
Finally, Sec.~\ref{sec:dataset} describes the construction of VideoEffect-130K.

\subsection{Problem Definition}
\label{sec:problem}

We study \emph{streaming video effect generation}, a human-centric video-to-video editing task that adds expressive effects to an incoming video stream under strict real-time constraints. 
The effects considered in this work include global or local stylization, accessories, headwear, makeup, decorative objects, and other effects commonly used in live commerce, entertainment, and creator scenarios. 
The task requires not only visually appealing effect rendering, but also strong preservation of human identity, body structure, background content, and temporal consistency.

Let \(\mathbf{X}=\{x_t\}_{t=1}^{T}\) denote a source video and \(\mathbf{Y}=\{y_t\}_{t=1}^{T}\) denote the target video after applying the desired effect. 
We introduce a reference image \(\mathbf{r}\) as an image-level effect condition, with a text prompt $\mathbf{c}_\text{txt}$ defining the desired effect.
Our goal is to learn a streaming generator \(G\) that maps the source video and reference condition to the edited video:
\begin{equation}
\hat{\mathbf{Y}} = G(\mathbf{X}, \mathbf{r}, \mathbf{c}_\text{txt}).
\end{equation}
For streaming inference, we divide the video into temporal chunks, \(\mathbf{X}=\{\mathbf{X}_0,\ldots,\mathbf{X}_M\}\) and \(\mathbf{Y}=\{\mathbf{Y}_0,\ldots,\mathbf{Y}_M\}\), and generate each target chunk causally:
\begin{equation}
\hat{\mathbf{Y}}_i = G(\mathbf{X}_{\leq i}, \hat{\mathbf{Y}}_{<i}, \mathbf{r}, \mathbf{c}_\text{txt}), \quad i=0,\ldots,M.
\end{equation}
This formulation allows the model to edit the video online while reusing previously generated chunks as temporal context.
At deployment, past predictions enter $G$ through
cached keys and values (Sec.~\ref{sec:distillation}), and a sliding
window bounds per-chunk cost over long streams. Together with chunk size
and few-step denoising, this gives a constant per-chunk latency budget.

\subsection{Reference-Conditioned In-Context Video Editing}
\label{sec:architecture}

StreamingEffect is built on a pretrained Wan2.2-TI2V-5B backbone~\cite{wan2025wan} and adopts a reference-conditioned in-context video editing architecture. 
Instead of injecting the source video through a separate control branch, we encode the source video, target video, and reference image into a shared latent token space and organize them into one context sequence. 
This design allows the denoising transformer to directly model the relationship among the original video content, the reference effect, and the target edited frames, which is important for preserving identity, background, and scene structure while adding new effects.

Let \(\mathbf{z}^{x}\), \(\mathbf{z}^{y}\), and \(\mathbf{z}^{r}\) denote the latent tokens of the source video, target video, and reference image. 
We arrange them into an interleaved sequence:
\begin{equation}
[\mathbf{z}^{r} \mid \mathbf{z}^{x}_{1} \mid \mathbf{z}^{y}_{1} \mid \mathbf{z}^{x}_{2} \mid \mathbf{z}^{y}_{2} \mid \cdots \mid \mathbf{z}^{x}_{M} \mid \mathbf{z}^{y}_{M}].
\label{eq:layout}
\end{equation}

Each pair \((\mathbf{z}^{x}_{i}, \mathbf{z}^{y}_{i})\) forms a temporal super-chunk $\mathcal{S}_i$ . 
A bidirectional model uses full attention over the sequence for high-quality short-clip editing, while the final streaming student replaces full attention with block-causal attention and KV caching for online generation.

The reference image \(\mathbf{r}\) provides keyframe-level effect control. 
During training we sample $\mathbf{r}$ uniformly at random from the frames
of $\mathbf{Y}$, forcing the model to learn how a single edited keyframe should guide effect rendering and temporal propagation. 
During inference, \(\mathbf{r}\) can be obtained by applying a strong image editing model or commercial image-editing API to a selected source frame. 
StreamingEffect then propagates this high-quality reference effect to subsequent frames in a temporally coherent manner. 
This design allows our streaming video editor to benefit from the strong visual quality of modern editing models while maintaining video-level consistency and low latency.

\subsection{Bidirectional Teacher Training}
\label{sec:teacher}

We first train a bidirectional teacher model $F_\phi$ using the in-context formulation above. 
The teacher receives the source video, the target video, and the reference image condition, and is optimized for visual quality and temporal coherence without enforcing streaming causality. 
Given a training tuple \((\mathbf{X},\mathbf{Y},\mathbf{r})\), the source video, target effect video, and sampled reference frame are encoded into latent tokens and processed by a transformer denoiser with full bidirectional attention. 
This allows each target token to attend to both past and future context, which is beneficial for effects that depend on face motion, body pose evolution, delayed appearance changes, or long-range temporal consistency.

We train the teacher with the native rectified-flow objective~\cite{esser2024scaling} of the backbone. 
Let \(\mathbf{z}^{y}\) be the clean target latent and \(\mathbf{z}^{y}_{t}\) its noisy interpolation at time \(t\). 
The teacher \(F_{\phi}\) predicts the flow direction conditioned on the source and reference latents:
\begin{equation}
\mathcal{L}_{\text{teacher}}
=
\mathbb{E}
\left[
\left\|
F_{\phi}(\mathbf{z}^{y}_{t}, \mathbf{z}^{x}, \mathbf{z}^{r},
\mathbf{c}_{\text{txt}}, t)
-
\mathbf{v}^{\star}
\right\|_{2}^{2}
\right],
\end{equation}
where $\mathbf{v}^{\star}$ denotes the target flow field.
We apply condition dropout during training on the semantic conditions $(\mathbf{z}^{r}, \mathbf{c}_{\text{txt}})$, encouraging the teacher to learn each as an independent signal.
The reference image is dropped at rate
$0.5$ and the text at rate $0.1$. 
When dropped, $\mathbf{z}^r$ is
replaced with an all-zero latent of identical shape.
This further enables classifier-free
guidance~\citep{ho2022classifierfreediffusionguidance} at inference. 
The trained teacher provides strong effect fidelity and temporal coherence, but it relies on full-sequence attention and many denoising steps, making it unsuitable for real-time streaming deployment.

\subsection{Two-Stage Distillation for Streaming Inference}
\label{sec:distillation}

Converting the bidirectional teacher into a real-time streaming editor requires more than swapping full attention with causal attention. 
The student must handle heterogeneous noise levels across time, as past chunks become clean and cached while the current chunk is still being denoised. 
Moreover, it must be trained on its own rollout distribution to reduce error accumulation over long streams.
We therefore use a two-stage distillation pipeline: Stage 1 performs causal adaptation with diffusion forcing~\cite{chen2024diffusionforcingnexttokenprediction}, and Stage 2 performs on-policy self-forcing~\cite{huang2025selfforcing} for few-step streaming generation to close the exposure-bias gap.

\paragraph{Stage 1: Bidirectional-to-Causal Adaptation.}
The first stage converts the bidirectional teacher into a causal autoregressive student. 
We keep the same interleaved super-chunk layout of Eq.~\eqref{eq:layout} but replace full attention with a block-causal mask: each super-chunk $\mathcal{S}_i$ attends bidirectionally within itself and to the reference $\mathbf{z}^r$ and all previous super-chunks
$\mathcal{S}_{j\le i}$, but not to future chunks.
At inference time, a persistent KV cache stores keys and values from previous chunks, avoiding repeated computation and enabling low-latency streaming generation.

Stage 1 also adapts the timestep distribution. 
Standard diffusion training samples a single timestep for the whole video, while streaming inference contains mixed noise states: previous chunks are already clean in the cache, whereas the current chunk remains noisy. 
We therefore assign each super-chunk an independent timestep $t_i$, including the clean state $t_i{=}0$, so the student observes heterogeneous noise configurations encountered during autoregressive generation. 
The student is initialized from the teacher ($G_\theta\!\leftarrow\!F_\phi$) and optimized with the same rectified-flow denoising objective under the causal mask. 
This stage makes chunk-wise streaming generation possible, but it still relies on teacher-forced inputs derived from ground-truth latents and therefore does not fully remove rollout-induced exposure bias.

\paragraph{Stage 2: On-Policy Self-Forcing for Few-Step Generation.}
The second stage closes the few-step gap and the exposure-bias gap simultaneously. 
Following Self Forcing~\cite{huang2025selfforcing}, we roll out
the Stage-1 student exactly as it will run at inference: across
chunks autoregressively, with each new chunk attending to a KV cache
$\mathrm{KV}_{<i}$ populated by the student's own previous-chunk
predictions $\hat{\mathbf{z}}^y_{<i}$. 
Within each chunk, we simulate the deployment-time $K$-step rectified-flow sampler under classifier-free guidance:  at each rollout step $k$ for chunk $i$, the student performs
two forward passes that share the same cache.
A conditional one
$\mathbf{v}_{\text{cond}} = G_\theta(\mathbf{z}^y_{i,t_k}, \mathbf{z}^x, \mathbf{z}^r, \mathbf{c}_{\text{txt}}; \mathrm{KV}_{<i}, t_k)$ and one
with both semantic conditions ablated $\mathbf{v}_{\text{uncond}} =
G_\theta(\mathbf{z}^y_{i,t_k}, \mathbf{z}^x, \varnothing, \varnothing; \mathrm{KV}_{<i}, t_k)$.
The two are blended at the teacher's CFG scale $w$
to produce $\tilde{\mathbf{v}}_{t_k} = (1{+}w)\mathbf{v}_{\text{cond}}
- w\mathbf{v}_{\text{uncond}}$ and the $\mathbf{x}_0$-prediction
$\hat{\mathbf{z}}^y_{i,0,k}$ that drives the rollout to step $k{+}1$.
Gradients flow only through the conditional forward and the unconditional
forward is detached.

Because our task is conditional video-to-video editing with paired data,
each $(\mathbf{X},\mathbf{r},\mathbf{c}_{\text{txt}})$ corresponds to a specific target $\mathbf{Y}$. 
This allows us to supervise the on-policy rollout directly with ground-truth target latents $\mathbf{z}^y$, rather than relying only on teacher scores, reward models, or discriminators~\cite{yin2024dmd, yin2024dmd2}. 
We use an SNR-weighted regression objective over the rollout:
\begin{equation}
\mathcal{L}_{\text{stage2}}
=
\frac{1}{\sum_k w_k}
\sum_{k=1}^{K}
w_k
\left\|
\hat{\mathbf{z}}^{y}_{0,k}
-
\mathbf{z}^{y}
\right\|_2^2,
\quad
w_k=\frac{(1-t_k)^2}{t_k^2}.
\end{equation}
SNR weighting emphasizes low-noise steps where the clean target is more strongly determined. 
This is key to stabilizing ultra-low-step generation and reducing error accumulation across chunks.

After Stage 2, the final student performs causal streaming inference with KV-cached attention and only a few denoising steps. 
In our implementation, the sampling process is reduced from 50 steps to 4 steps, enabling real-time 720p video effect generation while retaining much of the bidirectional teacher's visual quality and temporal coherence.

\begin{figure*}[!t]
\centering
\includegraphics[width=\linewidth]{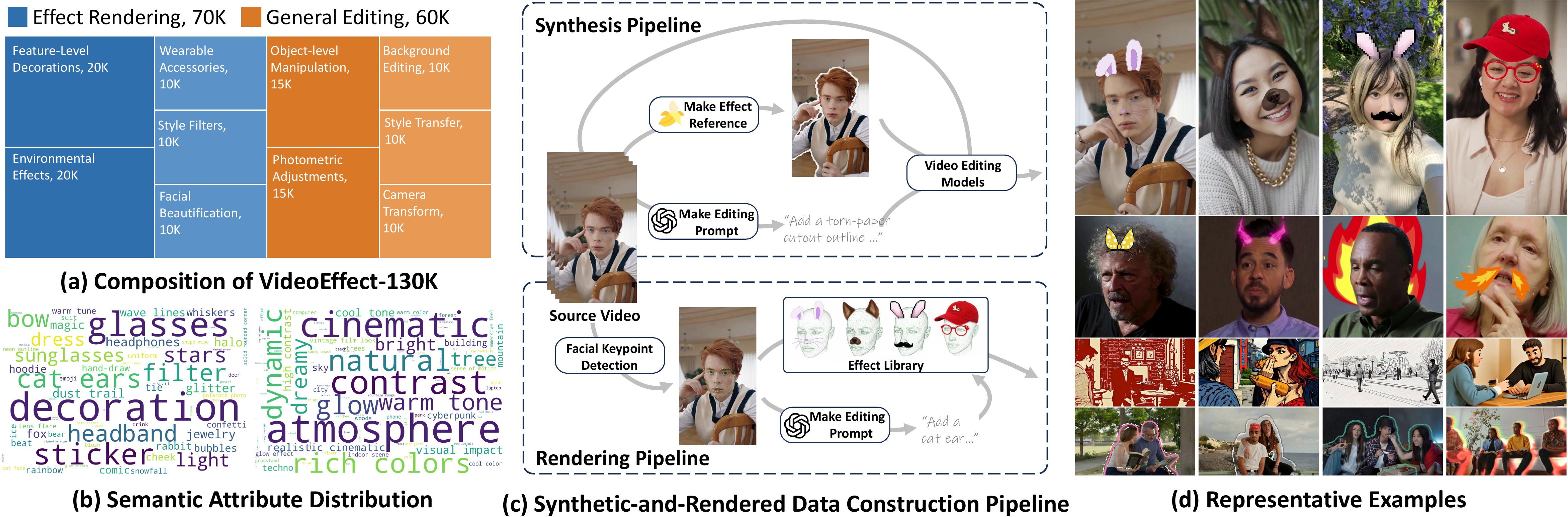}
\caption{\textbf{Construction and statistics of VideoEffect-130K.} The dataset contains 130K paired human-centric videos, including 70K rendered effect samples and 60K general editing samples across about 600 categories. It is built with a hybrid synthetic-and-rendered pipeline, and each sample consists of a source video, a reference effect image, and a target edited video.}
\label{fig:dataset}
\end{figure*}

\subsection{Dataset Construction}
\label{sec:dataset}

To support streaming video effect generation, we construct \textbf{VideoEffect-130K}, a large-scale paired video dataset for human-centric effect rendering and general video editing. As shown in Figure~\ref{fig:dataset}, the dataset contains about 130K source-target video pairs, including \textbf{70K effect-rendering samples} and \textbf{60K general-editing samples}. The effect-rendering subset covers feature-level decorations, environmental effects, wearable accessories, style filters, and facial beautification, while the general-editing subset includes object-level manipulation, photometric adjustments, background editing, style transfer, and camera transformation.

VideoEffect-130K is built through a hybrid synthetic-and-rendered pipeline. The \emph{synthesis} branch uses source videos, generated prompts, effect references, and video editing models to produce diverse target videos, improving coverage of open-ended editing instructions. The \emph{rendering} branch applies a platform-style effect library to source videos, optionally with facial keypoints and other human-centric cues for spatial alignment, providing stable supervision for common live-streaming effects such as facial accessories, headwear, makeup-like effects, local decorations, and atmosphere effects.

After construction, we apply filtering, trimming, categorization, and quality control to remove low-quality, temporally unstable, or poorly aligned pairs, and preprocess raw videos into 720p clips. Each sample is organized as a triplet of source video, reference image, and target video. During training, the reference image is randomly sampled from the target video, encouraging the model to learn reference-keyframe-conditioned effect rendering and temporal propagation over the video stream.

\begin{figure*}[!t]
\animategraphics[width=\linewidth]{5}{figs/result/frame_}{01}{09}
\caption{\textbf{Qualitative results of StreamingEffect.}
(a) \textbf{Image-guided:} propagating reference-image effects while preserving identity.
(b) \textbf{Text-guided:} following text prompts to generate diverse visual effects.
Readers can click and play video clips using {\color{red}\textbf{Adobe Acrobat}}.
}
\label{fig:result}
\end{figure*}

\begin{figure*}[!t]
\animategraphics[width=\linewidth]{5}{figs/compare4/frame_}{01}{09}
\caption{Qualitative comparison against open-source (Kiwi-Edit, OmniWeaving) and commercial (Kling O1, V3) baselines. Readers can click and play the video clips using {\color{red}\textbf{Adobe Acrobat}}.}
\label{fig:compare}
\end{figure*}

\begin{table}[t]
\centering
\caption{
Quantitative comparison under \textbf{image-conditioned} and \textbf{text-driven} settings. V-Cons., E-Ref., Text, and Quality are VLM scores for temporal consistency, effect-reference alignment, text-following, and overall quality. Lower is better for MSE/LPIPS; higher otherwise. Best in \textbf{bold}.
}
\label{tab:quant_image_condition}
\scriptsize
\setlength{\tabcolsep}{3pt}
\renewcommand{\arraystretch}{1.05}
\begin{tabular}{l ccccc cccc @{\hskip 8pt} ccc}
\toprule
\multirow{2}{*}{Method}
  & \multicolumn{9}{c}{Image-Guided}
  & \multicolumn{3}{c}{Text-Guided} \\
\cmidrule(lr){2-10} \cmidrule(lr){11-13}
& MSE$\downarrow$ & PSNR$\uparrow$ & SSIM$\uparrow$ & LPIPS$\downarrow$ & FPS$\uparrow$
& V-Cons.$\uparrow$ & E-Ref.$\uparrow$ & Text$\uparrow$ & Quality$\uparrow$
& V-Cons.$\uparrow$ & Text$\uparrow$ & Quality$\uparrow$ \\
\midrule
Kiwi-Edit
  & 0.0172 & 18.7 & 0.727 & 0.227 & \textbf{1.46} & 7.90 & 6.73 & 6.94 & 6.96
  & 7.72 & 7.12 & 6.81 \\
Kling O1
  & 0.0173 & 18.9 & \textbf{0.782} & 0.216 & -- & 8.10 & 8.18 & 8.35 & 7.82
  & 8.05 & 8.43 & 7.74 \\
Kling 3.0 Omni
  & 0.0184 & 18.5 & 0.723 & 0.240 & -- & 8.78 & 8.43 & 8.73 & 8.24
  & 8.61 & 8.62 & 8.15 \\
OmniWeaving
  & 0.0284 & 16.7 & 0.732 & 0.268 & 0.05 & 7.82 & 6.79 & 7.41 & 7.09
  & 7.65 & 7.58 & 6.94 \\
\textbf{Ours (Teacher)}
  & \textbf{0.0154} & \textbf{19.4} & 0.763 & \textbf{0.195} & 0.14
  & \textbf{8.98} & \textbf{9.02} & \textbf{8.75} & \textbf{8.41}
  & \textbf{8.84} & \textbf{8.68} & \textbf{8.19} \\
\bottomrule
\end{tabular}
\end{table}

\section{Experiments}

\subsection{Setup}

\paragraph{Implementation Details.}
StreamingEffect is built on the pretrained \textit{Wan2.2-TI2V-5B} backbone. All training and evaluation are performed at 720p with multi-aspect-ratio bucketed inputs ($1{:}1$, $9{:}16$, $4{:}3$, etc.), each clip containing $97$ pixel frames at $24$\,fps encoded into $25$ latent frames and grouped into $5$ super-chunks (chunk size $5$).
The bidirectional teacher is trained on $8$ H200 GPUs with LoRA, batch size $4$ per GPU, learning rate $1\times10^{-4}$, weight decay $1\times10^{-2}$, \textit{AdamW}, for $8{,}000$ steps.
\textbf{Stage~1 (Bidirectional$\to$Causal SFT)} initializes the causal AR student from the teacher and trains with FSDP, batch size $4$ per GPU, learning rate $5\times10^{-5}$, weight decay $1\times10^{-2}$, CFG scale $5.0$, for $3{,}000$ iterations; per-chunk independent timesteps are sampled uniformly in $[0.001, 0.999]$ to expose the student to heterogeneous noise configurations.
\textbf{Stage~2 (On-Policy Self-Forcing, 4-step)} starts from the Stage-1 checkpoint and runs on-policy rollouts with batch size $4$ per GPU, student learning rate $1\times10^{-6}$, weight decay $1\times10^{-2}$, for $3{,}000$ iterations, using a $4$-step shifted schedule with timestep list $[0.999, 0.937, 0.833, 0.624, 0.0]$.





\noindent\textbf{Baselines.}
We compare against two commercial systems, \textit{Kling O1} and \textit{Kling 3.0 Omni}~\citep{kling_omni_technical_report_2025}, accessed via public APIs, and two open-source video-editing models, \textit{Kiwi-Edit}~\citep{lin2026kiwi} and \textit{OmniWeaving}~\citep{pan2026omniweaving}, all evaluated under 720p, 97-frame configuration. None of these baselines support online streaming inference, so we compare against their offline outputs to assess effect quality independently of latency.

\noindent\textbf{Benchmarks.}
We evaluate on a held-out test split of $2{,}000$ samples drawn from the $70$K effect-rendering subset of VideoEffect-130K, covering around $600$ effect categories. 
Each test sample is a triplet of (source video, reference effect image, target edited video), with $97$-frame $720$p clips at multiple aspect ratios.

\noindent\textbf{Evaluation Metrics.}
We report full-reference metrics MSE, PSNR, SSIM, and LPIPS computed frame-by-frame between the prediction and the ground-truth target. FPS is measured on a single NVIDIA H200 GPU at $720{\times}1280$ resolution with each method's default sampling configuration, and is unavailable for the API-only commercial baselines; since FPS is determined by model architecture and sampling schedule rather than the conditioning modality, it does not change between image-conditioned and text-driven settings. Since effect generation is open-ended and many plausible edits satisfy the same instruction, we additionally use a VLM-based protocol in which \textit{Gemini~2.5~Pro}~\citep{comanici2025gemini} and \textit{GPT-4o}~\citep{openai2023gpt4} independently score each video on a $1$--$10$ scale along four dimensions averaged across the two models: \textbf{V-Cons.} (temporal consistency), \textbf{E-Ref.} (effect-reference alignment), \textbf{Text} (text-following), and \textbf{Quality} (overall editing quality, including identity and background preservation). 

\subsection{Qualitative and Quantitative Evaluation}

Figures~\ref{fig:result} and~\ref{fig:compare} show our qualitative results. Compared to StreamingEffect, the open-source baselines tend to lose human identity and exhibit flicker on faces and clothing, especially under motion. Commercial systems achieve stabler structure, but still suffer from identity drift and occasional loss of detailed reference effects. In contrast, our model consistently preserves identity, pose, and background while propagating the reference effect across the full clip with better temporal stability.


Table~\ref{tab:quant_image_condition} reports the main comparison under the image-conditioned setting. Our bidirectional teacher leads on full-reference fidelity, achieving better MSE, PSNR, and LPIPS than the baselines, and also obtains the best scores on all four VLM dimensions. These results suggest that the teacher can better preserve the target video structure while faithfully transferring the reference effect. The gains are most evident on effect-reference alignment and overall quality, indicating stronger reference-conditioned effect rendering and more stable human-centric editing. Open-source baselines lag behind the commercial systems on E-Ref.\ and Quality, showing that faithful reference-effect transfer to human-centric video remains an open challenge.

\subsection{Ablation Study}

\paragraph{Effect of streaming distillation.}
We ablate the two distillation stages in Table~\ref{tab:quant_stage_ablation}. The bidirectional teacher offers the strongest quality but its full attention and $50$-step sampling give well below real-time throughput at $720{\times}1280$. The Stage-1 causal AR student enables streaming via KV caching, yet naively reducing it to $4$ steps causes clear degradation across all VLM dimensions, confirming that ultra-low-step sampling is not free for a model trained only with multi-step targets.
Our final Stage-2 self-forcing student recovers most of this quality at $4$ steps while delivering an order-of-magnitude FPS gain over the teacher. This confirms that causalization enables online generation, while on-policy self-forcing is essential for making ultra-low-step streaming inference practical.

\begin{figure*}[!t]
\animategraphics[width=\linewidth]{5}{figs/ablation3/frame_}{01}{09}
\caption{\textbf{Qualitative ablation across distillation stages.}
\textbf{Bidirectional}: 50-step teacher with full attention. \textbf{Causal AR 4/50 steps}: Stage-1 causal student at 4 or 50 denoising steps. \textbf{Self-Forcing}: our final Stage-2 student (4 steps). Readers can click and play the video clips using {\color{red}\textbf{Adobe Acrobat}}.}
\label{fig:ablation}
\end{figure*}

\begin{table}[t]
\centering
\caption{
Ablation of the streaming distillation pipeline at $720{\times}1280$ on a single NVIDIA H200. Lower is better for MSE/LPIPS; higher otherwise. Best in \textbf{bold}.
}
\label{tab:quant_stage_ablation}
\scriptsize
\setlength{\tabcolsep}{2pt}
\renewcommand{\arraystretch}{1.05}
\begin{tabular}{l c ccccccccc @{\hskip 8pt} ccc}
\toprule
\multirow{2}{*}{Method} & \multirow{2}{*}{Steps}
  & \multicolumn{9}{c}{Image-Conditioned}
  & \multicolumn{3}{c}{Text-Driven} \\
\cmidrule(lr){3-11} \cmidrule(lr){12-14}
& & MSE$\downarrow$ & PSNR$\uparrow$ & SSIM$\uparrow$ & LPIPS$\downarrow$ & FPS$\uparrow$
  & V-Cons.$\uparrow$ & E-Ref.$\uparrow$ & Text$\uparrow$ & Quality$\uparrow$
  & V-Cons.$\uparrow$ & Text$\uparrow$ & Quality$\uparrow$ \\
\midrule
Bidirectional Teacher
  & 50 & \textbf{0.0154} & \textbf{19.4} & \textbf{0.763} & \textbf{0.195} & 0.14
       & \textbf{8.98} & \textbf{9.02} & \textbf{8.75} & \textbf{8.41}
       & \textbf{8.84} & \textbf{8.68} & \textbf{8.19} \\
Causal AR Student
  & 50 & 0.0261 & 16.6 & 0.678 & 0.276 & 0.63
       & 8.10 & 7.82 & 8.16 & 7.71
       & 7.98 & 8.05 & 7.52 \\
Causal AR Student
  &  4 & 0.0263 & 16.6 & 0.659 & 0.278 & 7.52
       & 7.93 & 7.47 & 7.63 & 7.26
       & 7.81 & 7.49 & 7.08 \\
\textbf{StreamingEffect}
  &  4 & 0.0251 & 16.7 & 0.666 & 0.269 & \textbf{14.1}
       & 8.37 & 8.14 & 8.55 & 7.75
       & 8.26 & 8.38 & 7.58 \\
\bottomrule
\end{tabular}
\end{table}

\section{Conclusion}

We presented \textbf{StreamingEffect}, an efficient real-time framework for human-centric streaming video effect generation. 
The task aims to add expressive and controllable effects to live video streams while preserving identity, background content, and temporal consistency under strict latency constraints. 
Our framework combines reference-conditioned in-context video editing with a two-stage distillation pipeline, converting a high-quality bidirectional teacher into a causal few-step streaming student. 
We further introduced keyframe control, which allows strong image editing models to provide high-quality reference effects that are propagated online. 
To support this task, we constructed \textbf{VideoEffect-130K}, a large-scale paired dataset that combines rendered effect data with synthetic/general-editing data for effect-centric human video editing.
Experiments show that StreamingEffect enables real-time, high-quality 720p video editing with strong visual fidelity on a single H200 GPU, suggesting a practical direction for deployable generative video effects.






\small
\bibliographystyle{plainnat} 
\bibliography{main}


\appendix

\section*{Appendix}

\section{Limitations}
\label{sec:limitations}

A primary limitation of our work stems from the composition of VideoEffect-130K.
Both the platform-rendered subset and the synthetic-editing subset are sourced
from short-video and live-streaming platforms whose creator content is
overwhelmingly human-centric, so the dataset is dominated by face, body, and
portrait scenes. As a consequence, StreamingEffect generalizes well to
human-centric effects---accessories, makeup, headwear, portrait stylization,
and atmosphere overlays---but is noticeably weaker on non-human scenarios such
as pure object close-ups, animals, landscapes, and product shots, where it can
miss the intended effect or apply it to the wrong region. Extending the
dataset with non-human effect categories, and training a more general
streaming editor, is a natural direction for future work.

\section{Evaluation Details}
\label{sec:suppl_llm}

We employ two strong multimodal large language models, Gemini~2.5~Pro~\citep{comanici2025gemini} and GPT-4o~\citep{openai2023gpt4}, as automated perceptual judges for human-centric video effect generation. To reduce single-judge bias and scoring variance, each metric is queried three times per model to obtain a median score, and the final reported score is the average of the medians from both Gemini~2.5~Pro and GPT-4o.

\textbf{Inputs.}
For every test sample, the judges receive (i)~a horizontally concatenated video with three columns---left: the input/source video, middle: the generated video, right: the ground-truth target video---(ii)~a reference effect image, and (iii)~a text prompt describing the desired effect. The judges are explicitly instructed to score only the generated video in the middle column, while using the source video, the ground-truth target video, the reference image, and the text prompt as evaluation context. They are asked to return JSON of the form \texttt{\{``score'': <1--10>, ``reason'': ``<sentence>''\}} for each of four metrics: \textbf{Video Temporal Consistency} (\texttt{V-Cons.}), \textbf{Effect--Reference Consistency} (\texttt{E-Ref.}), \textbf{Overall Editing Quality} (\texttt{Quality}), and \textbf{Text-Following Ability} (\texttt{Text}). Aggregated results are reported in Table~\ref{tab:quant_image_condition} and Table~\ref{tab:quant_stage_ablation} of the main paper.

\paragraph{Video Temporal Consistency (\texttt{V-Cons.}).}
The judges evaluate whether the generated video is temporally stable across frames. Criteria include the absence of flickering and jitter, no sudden appearance or disappearance of effects, stable identity throughout the clip, consistent background, natural and continuous motion, and the absence of temporal artifacts. Score~10 indicates a highly stable and coherent result over time; 8--9 minor temporal artifacts; 6--7 noticeable flicker or jitter; 4--5 strong instability; 1--3 severe flickering or temporal breakdown.

\paragraph{Effect--Reference Consistency (\texttt{E-Ref.}).}
The judges evaluate whether the generated effect matches the provided reference effect image. Criteria include the consistency of appearance, style, color, accessory, makeup, decoration, or visual effect with the reference image, and whether the effect is propagated correctly and stably over time. Score~10 indicates the generated effect is highly consistent with the reference; 8--9 minor deviation in style or color; 6--7 the effect is recognizable but partially mismatched; 4--5 only loosely related to the reference; 1--3 the effect is missing, unrelated, or clearly inconsistent with the reference image.

\paragraph{Overall Editing Quality (\texttt{Quality}).}
The judges evaluate the overall visual quality of the edited video. Criteria include realism and sharpness, identity preservation of the subject, preservation of body structure, preservation of the background, natural blending of the effect with the underlying scene, the absence of artifacts, and overall aesthetic quality. Pixel-level identity to the ground truth is explicitly not required, as long as the intended edit is achieved with high perceptual quality. Score~10 indicates excellent editing quality with natural and clean results; 8--9 high quality with minor artifacts; 6--7 acceptable but with visible flaws; 4--5 noticeable distortion or blending issues; 1--3 poor visual quality, severe artifacts, identity distortion, or failed editing.

\paragraph{Text-Following Ability (\texttt{Text}).}
The judges evaluate whether the generated video follows the provided text prompt. Criteria include whether the requested effect type, object, style, location, and semantic instruction are correctly reflected in the generated video. Score~10 indicates the generated video fully follows the text prompt; 8--9 the prompt is largely followed with minor omissions; 6--7 the main intent is captured but secondary instructions are missed; 4--5 the prompt is only partially followed; 1--3 the generated video ignores or contradicts the text prompt.

\textbf{Scoring rules.}
The judges are instructed to be strict but fair, and to follow several explicit rules: focus only on the generated video in the middle column; use the source video to judge content preservation; treat the ground-truth target video as a reference for the intended edit rather than a pixel-level target; use the reference image to judge effect consistency; and use the text prompt to judge semantic instruction following. High scores are not given when the effect looks visually pleasing but the subject's identity, body structure, or background are altered incorrectly; conversely, low scores are not given solely for slight deviations from the ground truth when the intended effect is correct and visually plausible.



\end{document}